*Commentary*

# PDDL2.1 — The Art of the Possible? Commentary on Fox and Long

**Drew McDermott**                                                                 DREW.MCDERMOTT@YALE.EDU
*Dept of Computer Science, Yale University,*
*PO Box 208285, New Haven, CT 06520–8285*

## Abstract

PDDL2.1 was designed to push the envelope of what planning algorithms can do, and it has succeeded. It adds two important features: *durative actions*, which take time (and may have continuous effects); and *objective functions* for measuring the quality of plans. The concept of durative actions is flawed; and the treatment of their semantics reveals too strong an attachment to the way many contemporary planners work. Future PDDL innovators should focus on producing a clean semantics for additions to the language, and let planner implementers worry about coupling their algorithms to problems expressed in the latest version of the language.

All things considered, Fox and Long have done a terrific job producing PDDL2.1. I know from experience that getting a committee to agree on a language requires a delicate combination of diplomacy and decree. The language extensions that emerged from the 2002 competition are not exactly what anyone wanted, but apparently everyone can live with them. PDDL2.1 is serving as a sturdy basis for evaluating and comparing planning algorithms, which is the prime purpose of the language in the first place. It appears that for the next competition only minor extensions, and no revisions, will be necessary.

On top of their work negotiating the syntax of the language, Fox and Long also produced a semantics, on display in their paper, plus a more elaborate semantics for fully autonomous processes, which did not make it into PDDL2.1, unfortunately. (I was on the 2002 competition committee, and, as I explain below, I am not as enthusiastic as others on the committee about the concept of "durative actions.")

Probably the most important innovation in PDDL2.1 is the introduction of objective functions for plans, thus making plan quality as important as plan existence. So far few planners have been able to do much with objective functions, which indicates how thoroughly we've all been conditioned by the classical-planning framework. Objective functions should become much more important in the future.

The main defect in PDDL2.1 is that its syntax and semantics are tailored too closely to a currently popular style of planner. For example, functions are allowed, but of exactly one kind, namely, those that take non-numeric arguments and denote time-varying numeric quantities. That is, in (*f* `---args---`), each *arg* must be an identifier and the overall value must be a number that can change from situation to situation. A paradigmatic example is (`amount-in tank1`), which might denote the volume of fuel in `tank1`. A term such as (`object-at-distance 3`) is not allowed. Why these restrictions? Because many planners eliminate all variables at the outset of a solution attempt by instantiating terms with all possible combinations of the objects mentioned in the problem statement. This tactic may sound unpromising, but for many problems of a reasonable size it works surprisingly well. However, as soon as the universe of objects becomes infinite the tactic stops working, and that means numbers can't be treated like ordinary objects. It also means that general functions can't be part of the language. If we had a function `midpoint: Location × Location ⟶ Location`, then it would generate an infinite set of terms such as `(midpoint loc-a (midpoint loc-b loc-a))`.

In PDDL 1.0, such problems did not arise because there were no functions in the language. The main goal in designing it was to agree on a lowest-common-denominator notation that many planners could obviously cope with, so that it could become a standard for problem statement. The language succeeded quite well in that regard,[1] which is why it is also the standard framework for discussions about where to go next. In those discussions, there are several relevant considerations:

1. What real-world problems need to be solved?

2. What problems lie just beyond the solvable fringe of the current state of the art?

3. What constructs can be given a clean semantics? Or any coherent semantics at all?

4. What constructs can current planning algorithms cope with?

I have listed these in declining order of importance, although I grant that they are all important. I believe that PDDL2.1 gives too much weight to consideration 4, and the example of functions is a good case in point. Functions can play several different roles in a logical theory, which is what PDDL domains are, when you get down to it. In an assembly-planning domain, someone might want a function `top` such that `(top cylinder-3)` denotes the top of a piece being worked on. How do the considerations above come into play?

1. Assembly planning is a real-world problem.

2. It lies well beyond what is currently solvable, probably too far beyond.

3. The semantics of functions in mathematical logic are well understood, and we can use the same solutions here.

4. Current planning algorithms can't cope with all functions, but they can easily be extended to handle functions like `top`, which can't be recursively nested.

Given these answers, and considering other examples, it seems clear that adding functions to PDDL is a good idea: it would make the language easier to use in realistic problems, and in many cases would impose a minimal burden on current planners. If the presence of functions makes some set of problems unsolvable by a planning system, then the system should detect when such a problem is encountered and go on to the next one. If we wanted to, we could add a `:functions` requirements flag to the language, but it hardly seems worth the trouble. But, as I said above, so much weight was attached to the abilities of current planners that PDDL2.1 ended up with a function declaration whose syntax of functions is needlessly restricted and whose semantics is needlessly complex.

---

1. To be precise, it succeeded well for action-based planners, and went nowhere for hierarchical planners.

We see the same phenomenon again with "durative actions," that is, actions that require a specified amount of time to execute. The committee had to thrash out a compromise about these things, mainly revolving around how far to go beyond the state of the art. A minority (including me, as well as Fox and Long) thought that the obvious next step was to be able to model autonomous processes, which differ from actions in two respects: they have continuous effects, and they run whenever their *conditions* are true ("precondition" is not quite the right term), no matter what the "target agent" (the one executing the plans) does. An example is boiling water: as long as there is water in a pot, and the water is at 100 degrees Celsius, the water will boil away, continuously decreasing in volume. The agent can make use of processes by making their conditions true or false at appropriate times.

Unfortunately, a majority of the committee thought putting processes into PDDL was too big a leap, and that we should instead add durative actions. As Fox and Long's paper shows, the term "durative action" really refers to two completely different species: actions that take a fixed amount of time no matter what, such as traveling from New York to London;[2] and actions whose duration is partly under the control of the planner, such as boiling water. The difference is flagged syntactically by whether the `:duration` field of a durative action is an equality (species 1) or an inequality (species 2). If an agent executes an action of species 1, it loses some of its "freedom" for the duration of the action. If the agent is sitting in an airplane, it's not out taking a hike. That seems unproblematic, but consider cleaning a warehouse, which might be modeled as taking an amount of time proportional to the "messiness" of the warehouse. It is a weird idealization to imagine that a robot might commit itself to cleaning the warehouse, and then essentially be a prisoner of this decision until the warehouse is clean.

Duratives of species 2 avoid this problem, by essentially sneaking autonomous processes into the theory in a strange form. We are allowed to use autonomous processes, just so long as we pretend the target agent is "executing" them. Rather than connect the process directly to its condition, we suppose that the agent can decide to stop the process at any point consistent with the constraints on `?d`, the duration of the action. So, in Figure 14 of Fox and Long's paper, rather than having an autonomous process that is started and stopped by changing the truth value of (`onHeatSource pot`), we say that (`onHeatSource pot`) becomes true or false when the agent starts or stops the `heat-water` action. No `turn-on` or `turn-off` actions are required.

The first remark to make is that the difference between an action the target agent can stop and one it must just wait to end should not be marked syntactically. Suppose it is possible for the agent to get locked out of the kitchen while it is boiling water. Then it can no longer stop the boiling. In PDDL2.1, it is impossible for this sort of thing to be expressed. The closest we can come is to make (`over all (in agent kitchen)`) be a condition of the durative action, but then as soon as the agent leaves the kitchen it must bring the `heat-water` action to a close, or its plan will be invalid.

My impression is that most planners that can handle duratives can handle only species 1, which is why the committee decided to include duratives. It seems clear to me that species 2 is headed for extinction in favor of straightforward autonomous processes.

---

2. Assuming flight time is fixed may seem too extreme an idealization, but allowing the time to vary (probabilistically?) would push PDDL far beyond its classical-planning roots; none of the controversies mentioned here ever question the knowability of the future.

Fox and Long define the semantics of duratives in terms of ordinary actions plus "monitoring actions" that make sure that conditions remain true over the intervals in which they're supposed to be true. It is possible to know exactly where these monitoring actions are supposed to be inserted because all changes in fluents are linear. This way of specifying semantics, has, unfortunately, just about run its course. As Fox and Long point out, future increases in the complexity of temporal constructs will make it harder to express the semantics of PDDL, and harder to verify that a plan is correct.

The details of durative semantics echo the issues that arose in connection with the semantics of functions. The tricky part about the semantics of actions is incorporating the *STRIPS assumption* that actions can be represented in terms of add lists and delete lists, which in turn requires assuming that situations can be represented as finite lists of atomic formulas. One might suppose that numbers would complicate this picture because there are an infinite number of them, but fortunately numbers in themselves don't compromise the STRIPS world view. If we specify a block's location in numerical coordinates, it still has only one location, and moving it involves deleting the assertion stating its old location and adding a new one.

Why, then, do Fox and Long work so hard to keep numerical assertions strictly separate from non-numeric? Why do they "flatten" action definitions before assigning them a semantics? Why are quantifiers handled by substituting all possible terms for the variables? The answer to all these questions is the same as for the odd restrictions on functions: Many current planners depend on generating all possible instances of an action.

It usually clarifies the semantics of a language greatly if it is defined without any direct connection to the implementation of a reasoning system for the language. In (McDermott, 2003) I sketch a formal semantics for an extension of PDDL containing true autonomous processes. The fulcrum of the framework is a set of truth conditions for process definitions. There is no obvious link to the requirements of a planning algorithm, and in fact the semantics allows processes that would be quite difficult to cope with or exploit. However, it is not hard to find subsets of process definitions, including those corresponding to durative actions, that current planners, with slight extensions, could handle.

One of the key goals of PDDL from the beginning has been to put pressure on the automated-planning community to make planners handle a more realistic class of planning problems. When new versions of PDDL are restricted in ways congenial to existing planners, it sends a mixed message, urging us into new territory, and at the same time reassuring us that our algorithms might still be basically correct. The planning community doesn't really need so much reassurance; we should opt for a domain-definition language with clear syntax and clean semantics and then find algorithms that can solve problems in the domains the language describes.